\documentclass
[%
	draftclsnofoot,  
	peerreviewca,  
]%
{IEEEtran}%

\IEEEoverridecommandlockouts
\usepackage{cite}
\usepackage{amsmath,amssymb,amsfonts}
\usepackage{graphicx}
\usepackage{textcomp}
\usepackage{xcolor}

\input{99_custommacros.sty}  
\newcommand{\numchannels}{N}
\newcommand{\channelid}{n}
\newcommand{\prob}{p}
\newcommand{\proboccupied}{\prob_{o}}
\newcommand{\probURLLC}{\prob_{p}}
\newcommand{\probURLLCcritical}{\prob_{p,\,c}}
\newcommand{\channelcoefficient}{\internalchannelcoefficient_{\channelid,\,\timeindex}}
\newcommand{\rayleighcoefficient}{|\tilde{\internalchannelcoefficient}|_{\channelid,\,\timeindex}}
\newcommand{\lengthoccupation}{l_{o}}

\newcommand{\return}{R_{\timeindex, \actioni}}
\newcommand{\reward}{\internalreward_{\timeindex}}
\newcommand{\rewardalt}{\internalreward_{\tau}}
\newcommand{\rewardsumcapacity}{\internalreward_{C,\,\timeindex}}
\newcommand{\rewardtimeoutURLLC}{\internalreward_{d,\,\timeindex}}
\newcommand{\rewardtimeoutURLLCcritical}{\internalreward_{d,\,c,\,\timeindex}}
\newcommand{\weightcapacity}{\internalweight_{C}}
\newcommand{\weighttimeout}{\internalweight_{d}}
\newcommand{\weighttimeoutcritical}{\internalweight_{d,\,c}}

\newcommand{\policy}{\pi}

\newcommand{\action}{\internalaction}
\newcommand{\actioni}{\internalaction_{i}}
\newcommand{\statet}{\mathbf{\internalstate}_{\timeindex}}
\newcommand{\statetplus}{\mathbf{\internalstate}_{\timeindex+1}}
\newcommand{\statesca}[1]{\internalstate_{#1,\, \timeindex}}

\newcommand{\loss}{\mathcal{L}}
\newcommand{\losstd}{\loss_{\text{TD}}}
\newcommand{\losslogprob}{\loss_{\text{LP}}}
\newcommand{\lossmaxentropy}{\loss_{ME}}
\newcommand{\weightlosslogprob}{\internalweight_{\text{LP}}}
\newcommand{\weightlossmaxentropy}{\internalweight_{\text{ME}}}

\newcommand{\networkparameters}{\boldsymbol{\internalparameter}}
\newcommand{\networkparameterst}{\boldsymbol{\internalparameter}_{\timeindex}}
\newcommand{\numparameters}{m}

\newcommand{\epsilongreedy}{\textsc{\small EG}\xspace}
\newcommand{\variancebased}{\textsc{\small VB}\xspace}
\newcommand{\maximumentropy}{\textsc{\small ME}\xspace}
\newcommand{\urllc}{\textsc{\small URLLC}\xspace}

\DeclareGraphicsExtensions{.pdf,.jpeg,.png}

\usepackage{tikz}
\usetikzlibrary
{%
	patterns,
	positioning,
	decorations.pathreplacing,
}%

\usepackage[hidelinks]{hyperref}
\usepackage[%
	noend  
]{algpseudocode}
\usepackage{siunitx}	
\usepackage{xspace}

\begin{document}
\addtolength{\floatsep}{-1mm}  
\addtolength{\textfloatsep}{-3mm}  
\addtolength{\abovedisplayskip}{-1mm}  

\title
{%
	On the Importance of Exploration\\for Real Life Learned Algorithms%
	\thanks{This work was partly funded by the German Ministry of Education and Research (BMBF) under grant 16KIS1028 (MOMENTUM).}%
	\thanks{This work was accepted for presentation at IEEE SPAWC 2022.}%
}%
\author{%
	\IEEEauthorblockN{%
		Steffen~Gracla%
		,
		Carsten~Bockelmann%
		\ and
		Armin~Dekorsy%
	}%
	\IEEEauthorblockA{%
		Dept. of Communications Engineering, University of Bremen, Bremen, Germany\\
		{Email: \{gracla, bockelmann, dekorsy\}@ant.uni-bremen.de}
	}%
}%
%
%
\maketitle
%


\begin{abstract}
	The quality of data driven learning algorithms scales significantly with the quality of data available.
	One of the most straight-forward ways to generate good data is to sample or explore the data source intelligently.
	Smart sampling can reduce the cost of gaining samples, reduce computation cost in learning, and enable the learning algorithm to adapt to unforeseen events.
	In this paper, we teach three Deep Q-Networks~(\dqn) with different exploration strategies to solve a problem of puncturing ongoing transmissions for \urllc messages.
	We demonstrate the efficiency of two adaptive exploration candidates, variance-based and Maximum Entropy-based exploration, compared to the standard, simple \(\epsilon\text{-greedy}\) exploration approach. 
\end{abstract}





\section{Introduction}
\label{sec:introduction}

In recent years the topic of machine learning~(\machinelearning) has reemerged with force due to breakthroughs in big data technology.
Data driven learning methods boast a profile of strengths that complements the more traditional model-based design well, being able to extract approximate solutions from empirical data where models fail.
As such, researchers from all fields of study are examining the feasibility of introducing \machinelearning into their areas of expertise, with promising results in communication technologies~\cite{dahrouj_overview_2021}.

While the early findings for \machinelearning in communication systems are encouraging, some challenges lie ahead in transferring the theoretical findings to real life applications.
Central to the strength of data driven technology is the quality of the learning data set itself; Although data collection has increased considerably in all areas of life, good real life data can remain costly and difficult to obtain.
A variety of approaches have been proposed to alleviate this issue.
For example, by generating data in a parameterized simulation environment, the data generation can be tuned to act as an inductive bias for optimal learning~\cite{kasgari_experienced_2020}.
However, it has been observed that such data can suffer from a model gap where the simulation environment does not capture real life in sufficient detail to transfer learnings~\cite{haarnoja_soft_2019}.
Further, approaches such as the Prioritized Experience Replay~\cite{schaul2015prioritized} aim to increase sample efficiency by extracting as much information as possible from the available data.
While this is valuable, these algorithms cannot increase the inherent quality of the data set.
If information required for successful learning is not captured by the data set, the algorithm cannot learn it.
Ultimately, the most straight-forward way to extract high quality samples from an environment is intelligent exploration.

This issue is of particular importance for the field of Reinforcement Learning~(\reinforcementlearning), where the algorithm generates its own data set as it learns.
A common, simple approach to help \reinforcementlearning-agents explore their environment is the \(\epsilon\)-greedy exploration, in which the agents make a random decision at a probability \( \epsilon \) instead of acting on their own.
The parameter \( \epsilon \) is annealed over the course of training as the agent gets ready to exploit what it has learned.
While better than no exploration, this simple mechanism comes with a set of drawbacks: 1)~In real life, high volumes of random actions can be costly; 2)~The exploration is likely to generate redundant samples, exploring actions that the agent is already confident about; 3)~If certain events are especially rare or only appear late during training, the random exploration probability \(\epsilon\) may have already been annealed too much to discover them.
Ideally, we would like to emulate curiosity during exploration, intelligently deciding to take a risk where uncertainty is high.

In this paper, we highlight some pitfalls of weak exploration.
We examine a medium access control problem where an agent is tasked with scheduling \urllc messages.
In order to do this, they may opt to puncture an existing transmission on orthogonal resource channels as proposed for 5G NR \urllc or wait to see whether a channel will become free in the near future.
We train three simple Deep~Q-Networks~(\dqn)~\cite{mnih2013playing} for this purpose: 1)~A \dqn with \( \epsilon \)-greedy exploration; 2)~A \dqn with stochastic output; And 3)~a \dqn with an approximate Maximum-Entropy-Learning~\cite{ziebart2008maximum} constraint that imposes a uniform prior onto the output, in effect pulling the agent away from committing to just one course of action.

In the following, we will first introduce the setup and notations of our \urllc puncturing problem and review the design of our \machinelearning agents.
We will then detail the differences in the employed exploration mechanisms and follow with practical examinations of their behavior in learning and in experiencing new situations.
We conclude by summarizing our findings.

\subsection{Related Work}
\label{sec:relatedwork}

The use of \machinelearning for QoS-optimal \urllc puncturing has been investigated by, \eg \cite{huang_deep-reinforcement-learning-based_2020, li_deep_2020}.
They show the general feasibility of using \machinelearning algorithms to learn efficient trade-off estimation in \urllc puncturing.
Research on smart exploration or data generation strategies has a long history. The authors in~\cite{amin_survey_2021} provide a rich survey of exploration mechanisms in the context of \reinforcementlearning.
More recently, methods such as SAC~\cite{haarnoja_soft_2019}, Munchhausen \dqn~\cite{vieillard_munchausen_2020}, and, in the context of communication technology, \cite{srivastava_parameterized_2021} have re-explored the concept of entropy-penalties to state-of-the-art \reinforcementlearning.
Our work brings the two topics together and highlights the importance of exploration in communication technology.
%




\section{Preliminaries}
\label{sec:preliminaries}
This paper assumes knowledge of stochastic gradient descent learning and feed-forward neural networks~(\neuralnetwork).
Matrices and vectors are denoted in boldface.

\section{Setup \& Notations}
\label{sec:setup}

In this section, we first describe mathematically the challenge of \urllc puncturing that we wish to learn a good strategy for, and secondly review the design and training of \dqn for decision making with deterministic or aleatoric output.

\subsection{URLLC Puncturing Simulation}
\label{sec:puncturing}
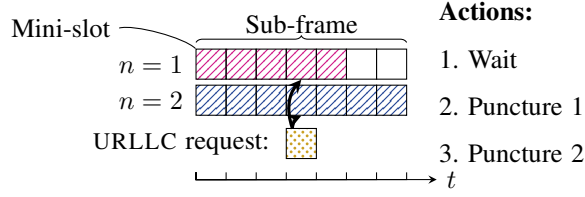
\begin{figure}[!t]
	\centering
	\begin{tikzpicture}
\tikzstyle{A0} = [-, >={stealth}, rounded corners]
\tikzstyle{A1} = [->, >={stealth}, rounded corners]
\tikzstyle{A2} = [<->, >={stealth}, rounded corners]
\tikzstyle{selfloop} = [looseness=4]
\tikzstyle{rb} = [draw, rectangle, minimum width=\rbside, minimum height=\rbside]
\tikzstyle{occupied1} = [pattern=north east lines, pattern color=ccolor1]
\tikzstyle{occupied2} = [pattern=north east lines, pattern color=ccolor2]
\tikzstyle{urllc} = [pattern = crosshatch dots, pattern color=ccolor4]

\newcommand{\rbside}{0.4cm}
\newcommand{\distancerby}{-1.2}

\node (textchan1)
	at (-2 * \rbside, 0 * \rbside)
	[]
	{\( \channelid = 1 \)};
	
\node (textchan2)
	at (-2 * \rbside, \distancerby * \rbside)
	[]
	{\( \channelid = 2 \)};

\node (rb11)
	at (0 * \rbside, 0 * \rbside)
	[rb, occupied1]
	{};
\node (rb12)
	at (1 * \rbside, 0 * \rbside)
	[rb, occupied1]
	{};
\node (rb13)
	at (2 * \rbside, 0 * \rbside)
	[rb, occupied1]
	{};
\node (rb14)
	at (3 * \rbside, 0 * \rbside)
	[rb, occupied1]
	{};
\node (rb15)
	at (4 * \rbside, 0 * \rbside)
	[rb, occupied1]
	{};
\node (rb16)
	at (5 * \rbside, 0 * \rbside)
	[rb]
	{};
\node (rb17)
	at (6 * \rbside, 0 * \rbside)
	[rb]
	{};

\node (rb21)
	at (0 * \rbside, \distancerby * \rbside)
	[rb, occupied2]
	{};
\node (rb22)
	at (1 * \rbside, \distancerby * \rbside)
	[rb, occupied2]
	{};
\node (rb23)
	at (2 * \rbside, \distancerby * \rbside)
	[rb, occupied2]
	{};
\node (rb24)
	at (3 * \rbside, \distancerby * \rbside)
	[rb, occupied2]
	{};
\node (rb25)
	at (4 * \rbside, \distancerby * \rbside)
	[rb, occupied2]
	{};
\node (rb26)
	at (5 * \rbside, \distancerby * \rbside)
	[rb, occupied2]
	{};
\node (rb27)
	at (6 * \rbside, \distancerby * \rbside)
	[rb, occupied2]
	{};

\draw [decorate, decoration={brace}]
	(-.5 * \rbside, -.6 * \distancerby * \rbside)
	--
	node [above] {Sub-frame}
	(6.5 * \rbside, -.6 * \distancerby * \rbside);

\draw []
	(-.5 * \rbside, 3 * \distancerby * \rbside)
	--
	(-.5 * \rbside, 3.2 * \distancerby * \rbside)
	--
	(0.5 * \rbside, 3.2 * \distancerby * \rbside)
	--
	(0.5 * \rbside, 3 * \distancerby * \rbside);
	
\draw []
(-.5 * \rbside, 3 * \distancerby * \rbside)
--
(-.5 * \rbside, 3.2 * \distancerby * \rbside)
--
(1.5 * \rbside, 3.2 * \distancerby * \rbside)
--
(1.5 * \rbside, 3 * \distancerby * \rbside);

\draw []
(-.5 * \rbside, 3 * \distancerby * \rbside)
--
(-.5 * \rbside, 3.2 * \distancerby * \rbside)
--
(2.5 * \rbside, 3.2 * \distancerby * \rbside)
--
(2.5 * \rbside, 3 * \distancerby * \rbside);

\draw []
(-.5 * \rbside, 3 * \distancerby * \rbside)
--
(-.5 * \rbside, 3.2 * \distancerby * \rbside)
--
(3.5 * \rbside, 3.2 * \distancerby * \rbside)
--
(3.5 * \rbside, 3 * \distancerby * \rbside);

\draw []
(-.5 * \rbside, 3 * \distancerby * \rbside)
--
(-.5 * \rbside, 3.2 * \distancerby * \rbside)
--
(4.5 * \rbside, 3.2 * \distancerby * \rbside)
--
(4.5 * \rbside, 3 * \distancerby * \rbside);

\draw []
(-.5 * \rbside, 3 * \distancerby * \rbside)
--
(-.5 * \rbside, 3.2 * \distancerby * \rbside)
--
(5.5 * \rbside, 3.2 * \distancerby * \rbside)
--
(5.5 * \rbside, 3 * \distancerby * \rbside);

\draw []
(-.5 * \rbside, 3 * \distancerby * \rbside)
--
(-.5 * \rbside, 3.2 * \distancerby * \rbside)
--
(6.5 * \rbside, 3.2 * \distancerby * \rbside)
--
(6.5 * \rbside, 3 * \distancerby * \rbside);

\draw [A1]
(-.5 * \rbside, 3.2 * \distancerby * \rbside)
--
(7.5 * \rbside, 3.2 * \distancerby * \rbside);

\node (timeindex)
	at (8 * \rbside, 3.2 * \distancerby * \rbside)
	[]
	{\(t\)};

\node (minislot)
	at (-5 * \rbside, -1 * \distancerby * \rbside)
	[]
	{Mini-slot};
	
\draw [A0]
	(minislot.east)
	edge [bend left, looseness = .5]
	(rb11.north west);

\node (urllcrequesttext)
	at (-1 * \rbside, 2.2 * \distancerby * \rbside)
	[]
	{\urllc request: };
	
\node (urllcrequest)
	at (3 * \rbside, 2.2 * \distancerby * \rbside)
	[rb, urllc]
	{};
	
\draw [A1, very thick]
	(urllcrequest)
	edge [bend left]
	(rb24.south);

\draw [A1, very thick]
	(urllcrequest)
	edge [bend left]
	(rb14.south);

\node (optonstext)
	at (10 * \rbside, 0.5 * \distancerby * \rbside)
	[align=left]
	{{\bfseries Actions:}\\
	1. Wait\\
	2. Puncture 1\\
	3. Puncture 2};

\end{tikzpicture}




	\caption{%
		URLLC Puncturing simulation for \( \numchannels = \num{2} \) Resources.
		Ongoing transmissions occupy at most one sub-frame of \( \num{7} \)~discrete blocks, or mini-slots.
		At a time step \( \timeindex \), there is a URLLC request.
		The puncturing agent can decide to wait, or puncture either of the mini-slots on the resources \( \channelid \).
	}
	\label{fig:puncturing_sim}
\end{figure}

\reffig{fig:puncturing_sim} schematically displays the puncturing simulation.
We assume our agent is a puncturing module that is part of a greater medium access protocol within a centralized communication traffic scheduler.
As typical in OFDM, transmissions are scheduled by the greater protocol on discrete sub-frames of fixed length.
Multiple orthogonal transmissions can be scheduled at the same time on \( \numchannels \)~available resources.
Each sub-frame is divided into \( \num{7} \) discrete blocks that we call mini-slots, akin to the numerologies specified in 5G NR.
At the beginning of each sub-frame, the greater protocol fills each available resource~\( \channelid \) with a probability~\( \proboccupied \) for a length of \(
{ \lengthoccupation \sim \text{U}(\num{5},\,\num{7})}
\) mini-slots drawn from a uniform random distribution.
Further, at the beginning of each sub-frame, a power gain \( \channelcoefficient = \rayleighcoefficient^{\num{2}} \) for each resource~\( \channelid \) with \( \rayleighcoefficient \sim \text{Rayleigh}(\num{1}) \) is drawn from a i.i.d. Rayleigh distribution for each resource.
This introduces a measure of stochasticity into the simulation environment.

Time~\( \timeindex \) moves discretely with the beginning of each mini-slot.
On every~\( \timeindex \), with a probability \(  \probURLLC \), a \urllc puncturing request may be posed to the puncturing agent.
Puncturing requests occupy one mini-slot and come in two types: 1)~The normal type has to be scheduled within the current sub-frame or else be discarded; 2)~The critical type has to be scheduled within the next mini-slot or else be discarded.
Only a low percentage~\( \probURLLCcritical \) of requests are critical.
The agent can then select one out of \( \numchannels + \num{1} \) options: Either schedule the request to one of the \( \numchannels \) available resource or do nothing.
If the agent decides to schedule to a resource~\( \channelid \), then any transmission ongoing in that resource is voided.

We therefore formulate three objectives for our puncturing agent:
\begin{enumerate}
	\item Do not interrupt ongoing transmissions unnecessarily;
	\item Puncture normal requests with as little influence on ongoing transmissions as possible;
	\item Puncture critical requests immediately.
\end{enumerate}
Mathematically, we formulate these objectives in a weighted reward sum, as is typical for \reinforcementlearning.
At the end of each mini-slot time step~\( \timeindex \), we determine three rewards~\( {(\rewardsumcapacity,\, \rewardtimeoutURLLC,\, \rewardtimeoutURLLCcritical) \in \numbersreal} \) for ongoing transmissions, normal requests and critical requests, respectively.
\begin{align}
	\rewardsumcapacity = \sum_{\channelid=1}^{\numchannels} \log (\num{1} + \channelcoefficient)
\end{align}
is the sum capacity achieved by ongoing transmissions within the mini-slot.
\begin{align}
	\rewardtimeoutURLLC &= 
	\begin{cases}
		\num{-1} 	& \text{if normal request discarded}\\
		\num{0}		& \text{else}
	\end{cases},\\
	\rewardtimeoutURLLCcritical &= 
	\begin{cases}
		\num{-1} 	& \text{if critical request discarded}\\
		\num{0}		& \text{else}
	\end{cases}
\end{align}
are indicator functions for whether the objectives 2)~and~3) were violated.
All three rewards are collected with their respective weights in the reward sum
\begin{align}
	\label{eq:reward}
	\reward =
		\weightcapacity\rewardsumcapacity
		+ \weighttimeout\rewardtimeoutURLLC
		+ \weighttimeoutcritical\rewardtimeoutURLLCcritical
\end{align}
for the learning agents to process.

\subsection{Deep Q-Networks}
\label{sec:dqn}

In order to select the most beneficial out of a number of options, one might consider the long term benefit of selecting each option.
In \reinforcementlearning, the long term benefit is typically defined using the immediate reward \(\reward\) as defined in \refeq{eq:reward} as
\begin{equation}
	\label{eq:returns}
	\return = \expectation_{\policy} \left[\sum_{\tau=\timeindex}^{\infty} \rewarddiscount^{\tau - \timeindex} \rewardalt | \action = \actioni \right]
\end{equation}
for an action~\( \actioni \), given that we follow some policy \( \policy \) afterwards.
Future rewards are typically scaled down exponentially by a discount factor \( \rewarddiscount \in [0,1] \) to devalue uncertain consequences in the far future.
If this long term benefit~\( \return \) is fully known, optimal decisions may be chosen by greedily selecting whichever action~\( \actioni \) has the highest~\( \return \) in each time step~\( \timeindex \).

\dqn attempt to approximate the long term benefit function~\refeq{eq:returns} via neural networks.
Feed-forward neural networks are parameterized mathematical functions with one or multiple inputs and outputs.
The relation between input and output can be influenced by tuning the \( \numparameters \)~network parameters~\( {\networkparameters \in \numbersreal^{\numparameters}} \).
The parameters~\( \networkparameters \) are tuned automatically, typically using variants of stochastic gradient descent, such that the optimal parameters minimize an objective function.
A \dqn \( \longtermrewards(\statet,\, \networkparameters) \) specifically will take as an input a vector~\( \statet \) that describes the current state of the system, to be described subsequently, and output an estimate for the long term benefit function~\( \return \) for all available actions~\( \actioni \).
When data points \( (\statet,\,\actioni,\,\reward,\,\statetplus) \) are experienced, the network parameters can be tuned to improve the estimate by minimizing the temporal difference error
\begin{align}
	\label{eq:losstd}
	\losstd = (&\longtermrewards(\statet,\, \actioni,\, \networkparameterst) \nonumber\\
	&- (\reward + \max_{i} \longtermrewards(\statetplus,\, \actioni,\, \networkparameterst)))^{\num{2}}.
\end{align}
This loss compares the networks current estimate \( {\longtermrewards(\statet, \actioni,\, \networkparameters)} \) with an updated estimate that incorporates the experienced immediate reward~\( \reward \) and following state~\( \statetplus \).

In this paper, we consider two variations on \dqn.
The first has the same number of \( {\numchannels + \num{1}} \) outputs as there are actions available, \ie the decision to either do nothing or puncture one of the \( \numchannels \) available resources.
The second type has twice the number of outputs, two for each action.
Using the well-known reparameterization trick, these pairs of outputs are interpreted as mean and log-standard deviation for a Normal distribution from which the output estimates are subsequently sampled.
This gives the second type of \dqn an inherent variance in decision making; The standard deviation may also be interpreted as a measure of uncertainty in the network output.

We summarize the system state~\( \statet \) in a simple vector with the entries
\begin{itemize}
	\item \(\statesca{1} \in [0, 1]\) is the current relative mini-slot position within the sub-frame;
	\item \(\statesca{2} \in \{0, 1\}\) is \( \num{1} \) if there is a puncturing prompt, else \( \num{0} \);
	\item \(\statesca{3} \in \{0, 1\}\) is \( \num{1} \) if there is a critical puncturing prompt, else \( \num{0} \);
	\item \(\statesca{4:4+\numchannels} \in [0, 1] \) is each resources current relative remaining occupation.
\end{itemize}
%




\section{Exploration Mechanisms}
\label{sec:explorationmechanisms}

In this paper, we compare three different learning agents.
All three use \dqn as described in the previous section, using three different exploration mechanisms: 1)~\(\epsilon\text{-greedy}\) exploration; 2)~Variance based exploration; 3)~Variance and approximate Maximum-Entropy based exploration.
This section will introduce their workings and differences.

\textit{1)~\(\epsilon\text{-greedy}\) exploration~(\epsilongreedy)} is using a \dqn with deterministic output.
At each time step~\( \timeindex \), this exploration mechanism has two options.
At a probability~\( \epsilon \), a random action is selected with uniform probability from the set \( \action \), \ie either do nothing or puncture a communication resource.
Alternatively, at a probability~\( {1 - \epsilon} \), the agent selects the action with the highest current \dqn long term benefit estimate, \( \max_{\actioni} \longtermrewards(\statet,\, \actioni,\, \networkparameterst) \).
Over the course of training, the probability~\( \epsilon \) is decayed, allowing the network to increasingly exploit the knowledge it has learned.

\textit{2)~Variance based exploration~(\variancebased)} is using a \dqn with stochastic output.
It introduces another term~\( \losslogprob \) to the loss function \refeq{eq:losstd} that is the sum of log probability densities~\(\text{lp}(\longtermrewards(\statet,\, \actioni,\, \networkparameterst))\) for the networks current sampled long term benefit estimates.
Therefore,
\begin{align}
	\loss &= \losstd + \weightlosslogprob\losslogprob, \\
	\text{with }\losslogprob &= \sum_{i=1}^{\numchannels+\num{1}} \text{lp}(\longtermrewards(\statet,\, \actioni,\, \networkparameterst)).
\end{align}
Parameter \( \weightlosslogprob \) can be tuned to scale the relative importance of each loss term.
Low variances lead to high log probability densities; Therefore, in order to minimize this new loss, the \dqn is enticed to keep variance high while still learning appropriate benefit estimates.

\textit{3)~Variance and approximate Maximum-Entropy based exploration~(\maximumentropy)} is using a \dqn with stochastic output.
Maximum Entropy Learning, known from algorithms such as Soft Actor-Critic~\cite{haarnoja_soft_2019}, seeks to encourage exploration by rewarding a learning agent for keeping their decisions at high entropy, \ie similar relative likelihood.
As a proxy, we apply the softmax function to the networks long term benefit estimates~\refeq{eq:losstd}, transforming the outputs into a pseudo-probability distribution to be able to calculate their entropy.
This implies that \( {\text{sm}_{i}(\longtermrewards(\statet,\, \action,\, \networkparameterst)) \in [\num{0}, \num{1}]} \) and \( {\sum_{i=\num{1}}^{\numchannels+\num{1}}\text{sm}_{i}(\longtermrewards(\statet,\, \action,\, \networkparameterst)) = \num{1} } \).
We calculate the entropy of these new terms and add it as a second term to the temporal difference loss function \refeq{eq:losstd},
\begin{align}
	\loss &= \losstd + \weightlossmaxentropy\lossmaxentropy, \\
	\label{eq:lossmaxentropy}
	\text{with }\lossmaxentropy &= \sum_{i=1}^{\numchannels+\num{1}} \log[\text{sm}_{i}(\longtermrewards(\statet,\, \action,\, \networkparameterst))].
\end{align}
Parameter \( \weightlossmaxentropy \) is used to tune the relative importance of the loss terms.
This new loss applies a uniform prior to the \dqn output, encouraging the network output to be similar in magnitude.
To promote the value of one action over the others, the parameter update needs to escape the pull of this new loss term. 
%




\section{Experiments}
\label{sec:experiments}

This chapter investigates the impact of the three different choices in exploration strategy, \(\epsilon\text{-greedy}\)~(\epsilongreedy), Variance based~(\variancebased), and approximate Maximum Entropy (\maximumentropy), detailed in the previous chapter.
We first iterate specific implementation details used in this paper, then evaluate the success of each strategy in terms of initial learning sample efficiency and ability to explore new, unseen events.
Finally, we break down the impact that these exploration strategies have on the overall puncturing performance.

\subsection{Implementation Details}
\label{sec:implementationdetails}

\begin{table}[!t]
	\renewcommand{\arraystretch}{1.3}
	\caption{Training Configuration}
	\label{tab:args}
	\centering
	\begin{tabular}{ll|ll}
		\hline
		Steps~\( \timeindex \) per episode  & \( \num{3000} \) & Episodes & \( \num{30} \) \\
		Resources~\( \numchannels \) & \( \num{2} \) & Rew. Discount~\( \rewarddiscount \) & \( \num{0.99} \) \\
		Prob. Tx~\( \proboccupied \) & \( \SI{70}{\percent} \) & Prob. \urllc~\( \probURLLC \) & \( \SI{10}{\percent} \) \\
		\(  \epsilon \) Initial & \( \num{0.99} \) & Episodes \( {\epsilon \rightarrow \num{0.0}} \) & \( \SI{50}{\percent} \)  \\
		Adam Learning Rate &  \( {1e-4} \) & Target Update & \( {1e-4} \) \\
		Hidden Layers \( \times \) Nodes & \(\num{2} \times \num{128}\) & VB Weight~\( \weightlosslogprob \) & \( {1e-2} \) \\
		Cap. Weight~\( \weightcapacity \) & \( \num{1} \) & \urllc Weight~\( \weighttimeout \) & \( \num{5} \) \\
		Critical Weight~\( \weighttimeoutcritical \) & \( \num{5} \) & ME Weight~\( \weightlossmaxentropy \) & \( e \) \\
		\hline
	\end{tabular}
\end{table}

The used \dqn are vanilla implementations with target networks~\cite{osband2016deep} as the only addition, which we found required to reach acceptable learning stability.
All simulation and \dqn parameters can be found in \reftab{tab:args}.
We select reward weights \(\weightcapacity, \weighttimeout\) assuming they have been tuned by an expert to reach the desired balance of \urllc time outs and transmission interruptions for a given application.
For the \epsilongreedy \dqn we opted for a linear decay of the exploration probability~\( \epsilon \) to zero after half of the training episodes.
For numerical stability, the softmax values in \refeq{eq:lossmaxentropy} are clipped to \( [\text{sm}(\cdot)]_{\num{1e-3}}^{1} \).
The Adam optimizer~\cite{kingma_adam_2015} is used to perform gradient descent updates.
We found most success using the penalized tanh activation function~\cite{hayou_impact_2019}.
We focus on a small number of \( \numchannels = \num{2} \) resources to restrict computational and design complexity, though the methods and conclusions presented are expected to scale to higher \( \numchannels \).
The design and tuning of stable \neuralnetwork learning for highly complex tasks represents its own challenge and would distract from the key research aim.

Simulation and learning are implemented in Python3.9 and TensorFlow.
For further implementation details, the full code implementation is available online in~\cite{gracla2022code}.

\subsection{Exploration Performance}
\label{sec:results}

\begin{figure}[!t]
	\centering
	\input{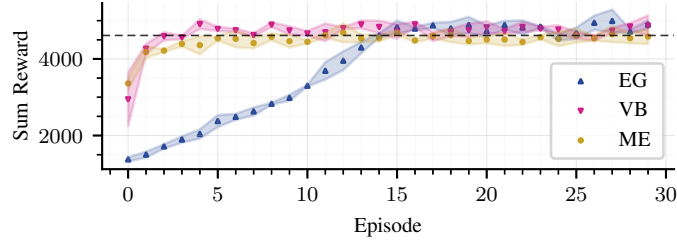}
	\caption{%
		The sum rewards achieved within a single training episode for \(\epsilon\text{-greedy}\)~(EG), Variance Based~(VB), and Maximum Entropy~(ME) Deep Q-Networks.
		The dashed black line shows the mean rewards achieved by manual scheduling.
		The EG DQN does not achieve strong results until its exploration parameter \(\epsilon\) is annealed to zero.
		Both other exploration candidates are adaptive and achieve competitive results much more quickly.
	}
	\label{fig:training_curve}
\end{figure}

For the initial training, we set the probability~\( \probURLLCcritical \) of encountering critical \urllc events to zero.
Three networks \epsilongreedy, \variancebased, \maximumentropy, with exploration strategies as described in \refsec{sec:explorationmechanisms}, are trained according to \reftab{tab:args}.
Training is repeated three times for each type, as the simulation and data generation have an inherent variance due to the simulation's stochastic setup.
The mean rewards achieved during training by each variant are depicted in \reffig{fig:training_curve}.
The dark dotted line denotes the mean rewards achieved by manual scheduling, which all three networks are able to meet eventually.
As expected, the \epsilongreedy \dqn generates many more samples to reach comparable performance to the other two candidates, as the exploration strategy is in no way adaptive.
This also results in many episodes with mediocre performance.

Next, we take the networks as trained in the previous step and confront them with a critical \urllc event in the first mini-slot of a sub-frame and both resources occupied.
Unlike normal \urllc events, critical events must be scheduled immediately or else time out.
We record the inferred results in \reftab{tab:reaction_critical}.
It shows the magnitude difference~(MD) in preference between action one compared to action two or three, calculated as
\begin{align}
	\longtermrewards(\statet,\, \action_{1},\, \networkparameterst) / ((\longtermrewards(\statet,\, \action_{2},\, \networkparameterst) + \longtermrewards(\statet,\, \action_{3},\, \networkparameterst)) / 2);
\end{align}
the log standard deviation of selecting action one; and the mean log standard deviation of selecting actions 2 and 3.
We note that all three \dqn show preference for action one, \ie doing nothing, therefore letting the critical \urllc request time out.
This is to be expected, as neither \dqn have seen a critical event during their training thus far and therefore cannot have learned the requirement to schedule it immediately.
In these results we can already spot the effects of our exploration strategies, where the \epsilongreedy \dqn has by far the largest difference in magnitude between the preferred action and the other actions and the \maximumentropy \dqn has the smallest difference.
Further, the \variancebased \dqn has considerably lower certainty in its decision to commit to doing nothing compared to the \maximumentropy \dqn.

\begin{table}[!t]
	\renewcommand{\arraystretch}{1.3}
	\caption{Mean (MD) and log std. (ls) reaction to new event}
	\label{tab:reaction_critical}
	\centering
	\begin{tabular}{%
		c|ccc
	}
		\hline
					&  \textbf{EG} 					&  \textbf{VB} 					&  \textbf{ME} \\
		\hline%
		\hline%
		MD 			& $80 \pm 0.03 \%$ 	& $6 \pm 0.05 \%$  	& $2 \pm 0.00 \%$ \\
		ls 1 		& \text{-}						& $-2.26 \pm 0.28$				& $-13.84 \pm 0.88$ \\
		ls 2 \& 3	& \text{-}						& $-0.27 \pm 0.09$				& $-13.96 \pm 1.18$ \\
	\end{tabular}
\end{table}

For the final step, we again load the pre-trained \dqn and again confront them with the new critical event situation from the previous step.
This time, we train them on this experience, confront them again, and repeat until a \dqn first decides to take a puncturing action, \ie an action other than action 1.
Due to the stochastic nature of the simulation we repeat this ten times for each pre-trained network.
\reftab{tab:training_steps_critical} displays the mean training steps required until each \dqn decides to explore this new situation.
On some training runs, the \epsilongreedy \dqn is stopped early after not deciding to explore for \( \num{10000} \) steps, while both other exploration strategies are able to start exploring massively more early in every training.

\begin{table}[!t]
	\renewcommand{\arraystretch}{1.3}
	\caption{Mean and std. training steps until exploring new event}
	\label{tab:training_steps_critical}
	\centering
	\begin{tabular}{ccc}
			\hline
			\bfseries EG & \bfseries VB & \bfseries ME \\
 			\hline\hline
			\(\num{6800} \pm \num{4525} \) & \(\num{96} \pm \num{17} \) & \(\num{22} \pm \num{2} \)\\
			\hline
		\end{tabular}
\end{table}

\subsection{Puncturing Performance Impact}
\begin{figure}[!t]
	\centering
	\input{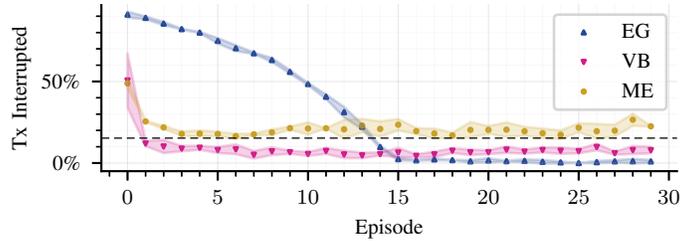}
	\caption{%
		Relative transmissions interrupted for puncturing for \(\epsilon\text{-greedy}\)~(EG), Variance Based~(VB), and Maximum Entropy~(ME) Deep Q-Networks over the course of training.
		The dashed line represents the mean result achieved by manual puncturing.
	}
	\label{fig:training_tx_interrupted}
\end{figure}

\begin{figure}[!t]
	\centering
	\input{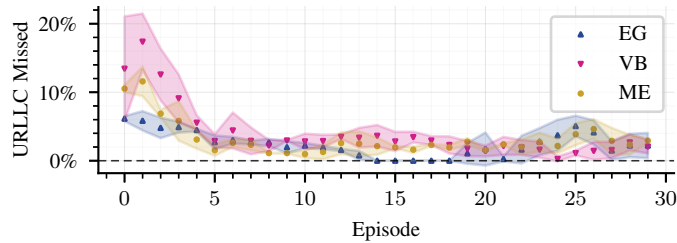}
	\caption{%
		Relative URLLC prompts missed over the course of training for \(\epsilon\text{-greedy}\)~(EG), Variance Based~(VB), and Maximum Entropy~(ME) Deep Q-Networks.
		The dashed line represents the mean result achieved by manual puncturing.
	}
	\label{fig:training_urllc_missed}
\end{figure}

After examining how the adaptive mechanisms have improved their respective networks' exploration strategies, we next examine their impact on asymptotic reward sum performance.
\reffig{fig:training_tx_interrupted} breaks down the ratio of transmissions interrupted by puncturing over the course of training, while \reffig{fig:training_urllc_missed} shows the ratio of \urllc prompts missed over the course of training.
Both \variancebased and \maximumentropy show slightly weaker asymptotic performance on the transmissions interrupted.
We attribute this to three factors:
\begin{enumerate}
	\item All networks converge to slightly different puncturing strategies with slightly different focus on each sub-task, for similar approximate overall performance on the optimization metric \( \reward \).
	For example, the manual puncturing, represented by the black dotted line, puts a heavy focus on catching \urllc requests, leading to a slightly increased amount of transmissions interrupted;
	\item Adding another term to the optimization function, in the form of exploration penalties, does represent a potential loss in optimality.
	The optimization focus is no longer to just optimize the target metric~\( \reward \), but to balance it with additional constraints.
	This effect is somewhat mitigated by \variancebased and \maximumentropy still being greedy schemes, \ie selecting the highest estimate action, which leads to accuracy on the non-maximum estimates being less important.
	Further, this effect is controllable via the weighting factors \( \weightlosslogprob, \weightlossmaxentropy \).
	In this paper, the weight for \maximumentropy in particular was set to a value that leads to excellent exploration performance for this application, as expressed in \reftab{tab:reaction_critical} and \reftab{tab:training_steps_critical};
	\item The training regimen used in this paper was not adjusted for each network's training and may therefore favor one variant over another.
\end{enumerate}

%


\section{Conclusions}
\label{sec:conclusions}

In this paper we examined an optimization problem in puncturing ongoing transmissions for \urllc messages of differing priority.
We implemented three learning agents, one with a standard \( \epsilon\text{-greedy} \) deterministic Deep Q-Network~(\dqn) and two \dqn with stochastic output.
For the stochastic \dqn we implemented exploration strategies based on a variance penalty and Maximum Entropy Learning, respectively.
Both stochastic \dqn exploration strategies encourage the learning agent to explore when uncertain and to not commit too heavily onto a single course of action.
While all three agents were able to learn to solve the optimization problem, we showed how the adaptive exploration strategies can lead to significant gains in learning sample efficiency and ability to adapt to unforeseen events, both of which we consider to be crucial for real life learned algorithms.
However, no exploration algorithm is universally optimal, and therefore must be applied mindful of their limitations.
\bibliographystyle{ref/IEEEtran}
{%
	\makeatletter  
	\clubpenalty10000  
	\@clubpenalty \clubpenalty
	\widowpenalty10000
	\makeatother
	
	\bibliography{ref/IEEEabrv,ref/references}
}%

\end{document}